%% file: main.tex
\documentclass[nonacm, sigconf]{acmart}

\usepackage{algorithm} 

\usepackage[noend]{algpseudocode}
\usepackage{graphicx}

\usepackage{subcaption}
\usepackage{caption}
\usepackage[utf8]{inputenc}
\usepackage{amsmath}
\usepackage[normalem]{ulem}
\usepackage{xcolor}
\usepackage{multirow}
\usepackage{booktabs}
\usepackage{enumitem}

\renewcommand\footnotetextcopyrightpermission[1]{}
\newcommand{\mypara}[1]{{\smallskip \noindent \bf #1}\hspace{0.1in}}

\newcommand{\myname}{{\bf CACTUS}}

\begin{document}

\title{Efficient IoT Inference via Context-Awareness}

\author{Mohammad Mehdi Rastikerdar}
\email{mrastikerdar@cs.umass.edu}
\affiliation{\institution{University of Massachusetts Amherst}
\country{United States}}

\author{Jin Huang}
\email{jinhuang@cs.umass.edu}
\affiliation{\institution{University of Massachusetts Amherst}\country{United States}}

\author{Shiwei Fang}
\email{shiweifang@cs.umass.edu}
\affiliation{\institution{University of Massachusetts Amherst}\country{United States}}

\author{Hui Guan}
\email{huiguan@cs.umass.edu}
\affiliation{\institution{University of Massachusetts Amherst}\country{United States}}

\author{Deepak Ganesan}
\email{dganesan@cs.umass.edu}
\affiliation{\institution{University of Massachusetts Amherst}\country{United States}}

\date{} 

\begin{abstract}
While existing strategies to execute deep learning-based classification on low-power platforms assume the models are trained on all classes of interest, this paper posits that adopting context-awareness i.e. narrowing down a classification task to the current deployment context consisting of only recent inference queries can substantially enhance performance in resource-constrained environments.
We propose a new paradigm, \myname{}, for scalable and efficient context-aware classification where a micro-classifier recognizes a small set of classes relevant to the current context and, when context change happens (e.g., a new class comes into the scene), rapidly switches to another suitable micro-classifier. 
\myname{} features several innovations, including optimizing the training cost of context-aware classifiers, enabling on-the-fly context-aware switching between classifiers, and balancing context switching costs and performance gains via simple yet effective switching policies. 
We show that \myname{} achieves significant benefits in accuracy, latency, and compute budget across a range of datasets and IoT platforms. 
\end{abstract}
\settopmatter{printfolios=true}

\maketitle

\input{Intro-V3}
\input{Case}
\input{DesignOverview}

\input{Design}
\input{Evaluation}
\input{Related}
\input{Conclusion}
\balance
\bibliographystyle{plain}

\end{document}

%% file: Intro-V3.tex
\section{Introduction}

There has been significant recent interest in executing deep learning models on low-power embedded systems and IoT devices (e.g. \cite{lai2018enable, warden2019tinyml,mazumder2021survey}). 
These devices are heavily resource-constrained in terms of compute and storage, and typically need to operate at very low energy budgets to ensure that they can run for weeks or months on battery power.

Prior approaches to squeeze deep learning models fall into three categories.
The first is approaches to compress the model via pruning, quantization, and knowledge distillation. 
For example, model pruning~\cite{han2015deep} and model quantization~\cite{jacob2018quantization, li2020train} techniques reduce the number of the neurons in the models or the number of the bits of the feature representation to achieve less compute.  
The second category is model partitioning, which executes the first few layers of a DNN on the IoT device and the remaining layers on the cloud~\cite{kang2017neurosurgeon,teerapittayanon2017distributed,choi2018deep, cohen2020lightweight, huang2020clio, itahara2021packet}.
The third category is the early-exit inference which takes advantage of the fact that most data input can be classified with far less work than the entire model. 
Hence, early exit models attach exits to early layers to execute the easy cases more efficiently~\cite{teerapittayanon2016branchynet, kaya2019shallow}.


\mypara{Context-aware inference.} 
We argue that one dimension that has not been fully explored in prior efforts is \emph{context-aware inference}. 
Context-aware inference narrows down a prediction task to the current context -- a much shorter time window than the overall time span of the deployed task -- to improve inference efficiency. 
Rather than attempting to squeeze the original deep learning model into memory, 
we instead load a smaller model that is relevant to the current context. 
We refer to the smaller model as a ``context-aware classifier'', noted as \textit{$\mu$Classifier}. 


In this work, we define ``context'' as a time window where only a subset of classes (called \textit{active classes}) are likely to appear.  
A context change indicates a transition to a time window whose active classes are different because a new class comes into the scene. 
We note that other definitions of contexts are possible, but not the focus of this paper. For example, a context could also refer to a particular difficulty level of inputs for all classes such as different weather patterns (rain, fog, snow) or light conditions (bright sunlight/overexposure). 
These definitions are orthogonal to context changes caused by variations in active classes which is our focus.

Context-aware inference is particularly effective in IoT settings because, in many applications, the context can remain fairly static for extended periods.
For instance, a trail camera positioned in a forest may primarily observe the same set of animals. Similarly, a doorbell camera in a residential area is likely to capture the same types of vehicles and individuals. 
Context-aware inference could be beneficial in various other applications such as industrial monitoring systems where machinery states change infrequently or in smart agriculture where monitoring conditions may remain constant over long growing seasons. This idea holds even greater significance in environments where compute and memory resources are scarce, such as in many IoT devices.

\mypara{Challenges: } Context-aware inference presents several new research challenges. During training, we need to efficiently create a $\mu$Classifier for every possible context, but the number of contexts grows exponentially with the number of classes. 
For example, if there are 100 classes and we consider 3-class $\mu$classifiers, there are $C^{3}_{100} = 161700$ $\mu$Classifers. 
In addition, the optimal configuration of a $\mu$Classifer depends on the similarity of classes in the context.  
A context whose classes have similar patterns would be harder to recognize and need to use a $\mu$Classifier that has a higher model capacity.  
If we consider 10 configurations per $\mu$Classifier, then nearly 1.6M models need to be trained to identify the optimal $\mu$Classifers for all possible contexts. 
Training $\mu$Classifiers with every configuration for each context cannot scale with a large number of configurations and classes.


At inference time, two additional challenges need to be tackled.  
First, we need a robust and lightweight method to detect context change so that we can switch to an appropriate $\mu$Classifier for the new context. 
Since context change detection lies on the critical path of every prediction, 
the detector needs to be very fast to avoid adding significant computation overhead to the $\mu$Classifier.

Second, after a new class is detected, we need to determine which $\mu$Classifier to switch to, i.e. what are the active classes in the new context. 
One option is to switch to a $\mu$Classifier that handles more classes, including the new class detected, to reduce the frequency of future context switches. 
However, a $\mu$Classifier with more classes tends to be more computation-intensive and thus has higher inference costs per input frame.
Another option is to use a $\mu$Classifier that recognizes only the new class. This can be more efficient but will cause more frequent context switches and thus higher switch costs. 
Thus, we need to design a context switching policy that considers both the number of active classes in the new context and the overhead due to context switches.


Existing approaches to explore context awareness do not address all dimensions of the problem.
The most relevant work ~\cite{palleon, han2016mcdnn} leverages class skewness in video processing to improve efficiency but rely on cloud servers to switch between models. This makes it less applicable to devices with limited network connectivity. 
Other approaches assume known class skews~\cite{FAST} or tailor DNN inference to specific applications without studying how to handle context changes effectively~\cite{kang2017noscope, jiang2018chameleon, leontiadis2021s}.

\mypara{The \myname{} approach:} 
We propose a novel paradigm for scalable and efficient context-aware classification that we refer to as \myname{}\footnote{CACTUS: Context-Aware Classification Through Ultrafast Switching.}.

Our work has three major contributions. 
First, we develop an \emph{inter-class similarity} metric to estimate the difficulty in classifying the active classes in each context.
We show that this metric is highly correlated with the optimal choice of configurations for the $\mu$Classifier, and hence allows us to rapidly and cheaply estimate the best configuration for each $\mu$Classifier without incurring training costs.

Second, we design an efficient context change adaptation pipeline by splitting the process into two parts -- a lightweight context change detector that executes on each image as a separate head attached to each $\mu$Classifer, and a more heavyweight context predictor that executes a regular all-class classifier to identify the new class. 
This separation allows us to invoke the heavyweight classifier sparingly thereby reducing the computational cost.

Third, we develop a simple yet effective context switch policy that adaptively determines what size $\mu$Classifer to switch to depending on how fast active classes change in a particular deployment environment. This allows us to achieve a balance between context switching costs and the performance gain of context-aware inference. Furthermore, our system can deal with variable IoT-Cloud connectivity to enable unattended operation where the device switches between popular $\mu$Classifers that are locally stored, as well as cloud-assisted operation where the device can request new $\mu$Classifers to be created and downloaded for new contexts.


We implement \myname{} by modifying EfficientNet-B0 \cite{efficientnet} and conduct extensive evaluation across five datasets (STL-10 \cite{STL10}, PLACES365 \cite{PLACES365}, and three Camera Trap datasets, Enonkishu \cite{Camera_trap_enonkishu}, Camdeboo \cite{Camera_trap_camdeboo}, and Serengeti (season 4) \cite{Camera_trap_serengeti}) and multiple IoT processors (Risc-V(GAP8) \cite{URL:GAP8} and low-end ARM 11 \cite{URL:Arm11} to high-end Cortex-A77 \cite{URL:Cortex-A77}). 

We show that:
\begin{itemize}[leftmargin=0.24in]
    \item \myname{} reduces computational cost by up to 13$\times$ while outperforming all-class EfficientNet-B0 in accuracy.
    \item Our inter-class similarity metric accurately captures the hardness of a set of classes in a $\mu$Classifier (0.86-0.97 Pearson's coefficient with classification accuracy).
    \item \myname{}'s context change detector has 10-20\% fewer false positives and false negatives than an alternate lightweight detector that uses similar FLOPs. 
    \item End-to-end results on the Raspberry Pi Zero W and GAP8 IoT processors show that \myname{} can achieve 1.6$\times$ -- 5.0$\times$ speed-up and up to $5.4\%$ gain in accuracy compared to All-Class EfficientNet. 
    \item We also show that \myname{} in Local-Only setting has up to 4.1$\times$ speed-up compared to All-Class EfficientNet by storing at most 30 $\mu$Classifiers.
    \item Our context switching policy leads to higher speedups in comparison to alternative policies (up to 2.5$\times$) while the accuracy is on a par with them.
    
\end{itemize}


%% file: Case.tex
\begin{figure*}[t]%
    \centering
    \subfloat[\centering Fewer classes is better\label{fig:accuracy_vs_classes}]{{\includegraphics[width=.32\textwidth]{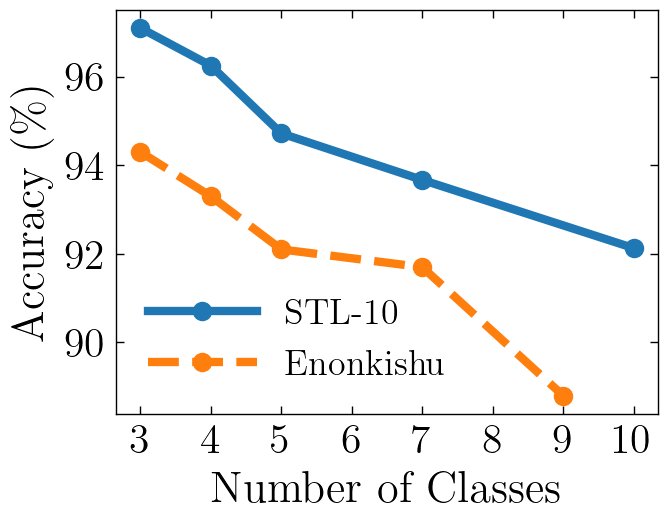}}}%
    \subfloat[\centering Pruning does not reduce gap (STL-10) \label{fig:accuracy_vs_size}]{{\includegraphics[width=.32\textwidth]{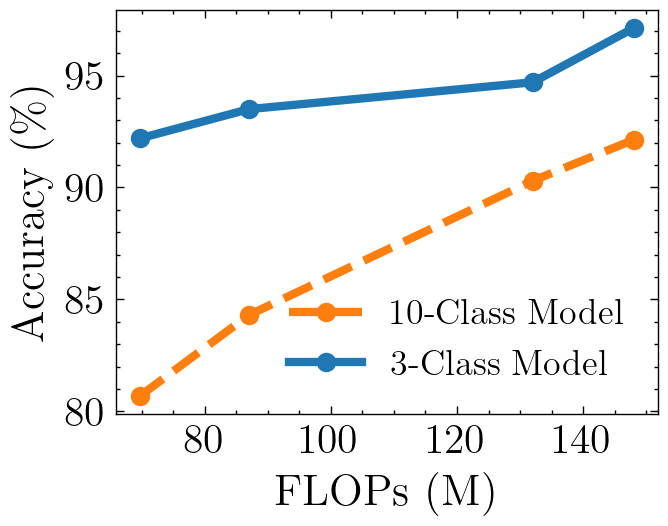}}}%
    \subfloat[\centering Early exit does not reduce gap \label{fig:early_exit accuracy_vs_size}]{{\includegraphics[width=.32\textwidth]{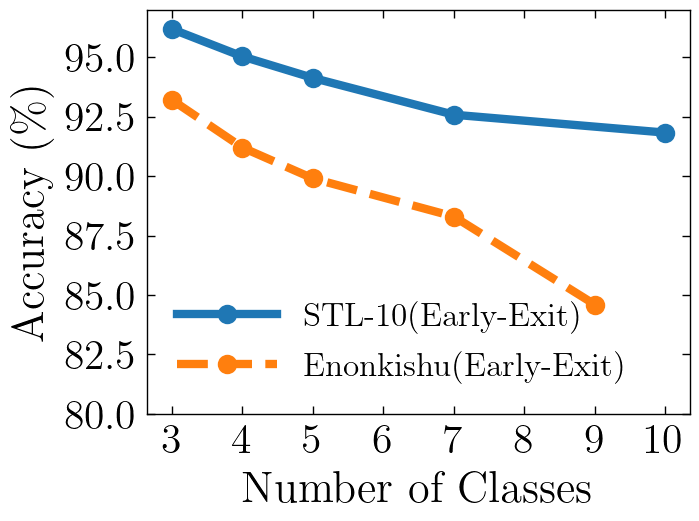}}}%
    \caption{Context-aware $\mu$Classifiers can provide substantial performance advantages. (a)  shows that accuracy drop is substantial as \#classes increases; (b) shows that model compression expands this gap further and (c) shows a similar trend for early-exit models.}
\label{knn_performance}
\end{figure*}

\section{Case for $\mu$Classifiers}
\label{sec:motivations}

In this section, we compare two designs for squeezing DNNs on IoT devices ---  the canonical method of training and compressing an \textit{all-class model} and our approach of using a \textit{$\mu$Classifier} that has fewer classes. 

\mypara{Observation 1: Using fewer classes has significant performance benefits.} 
Intuitively, if a classifier has fewer classes than another, it should be more accurate given the same resources. 
But how much is this gain in accuracy?

To answer this, we look at the accuracy of  EfficientNet-B0 on different subsets of classes from STL-10 \cite{STL10} and a Camera Trap dataset called Enonkishu ~\cite{Camera_trap_enonkishu} (See \S\ref{sec:settings} for details on the datasets). 
Figure~\ref{fig:accuracy_vs_classes} shows the average accuracy across all 3-class subsets of the dataset,  4-class subsets, and so on. 
Post-training dynamic range quantization~\cite{dynamicq} is applied to all models. 
We see that accuracy increases steadily as the number of classes decreases: the overall increase is by about 5\% and 5.5\% for STL-10 and Enonkishu respectively when we go from 10 classes to 3 classes. 
This indicates the substantial accuracy advantage of $\mu$Classifiers over an all-class model given the same DNN architecture. 

\mypara{Observation 2: Model compression does not diminish the advantage of $\mu$Classifiers.} 
In Figure~\ref{fig:accuracy_vs_classes}, we used EfficientNet-B0 with quantization but did not apply other techniques that are available for optimizing deep learning models for embedded MCUs (e.g. weight sparsification, filter pruning, and others \cite{he2018amc,liu2019metapruning,bhattacharya2016sparsification,lai2018cmsis,yao2018fastdeepiot,lane2015deepear,liu2018demand,cheng2017survey}). 
We now ask whether these techniques can bridge the accuracy gap between $\mu$Classifiers and the all-class model.

Figure~\ref{fig:accuracy_vs_size} shows the performance gap between a 10-class and 3-class model after model pruning and quantization. 
We apply filter pruning~\cite{str_p1} which reduces the number of filters in each layer based on $\ell_1$ norm. Each point represents a model architecture with a specific pruning level. The accuracy of each point on the dashed line that represents a 3-class model is computed by averaging the accuracy of all 3-class subsets. 
We see that the accuracy gap between the $\mu$Classifier and all-class classifier increases as resources reduce (5\% at 148 MFLOPs to 11.5\% at 70 MFLOPs).  
The result highlights that model optimization techniques cannot diminish the accuracy advantage of $\mu$Classifiers over the all-class model. 

\mypara{Observation 3: The benefits hold for early exit models.} 
Another approach to reduce the computational requirements of executing models on IoT devices is to use early exit, where the model only executes until it is sufficiently confident about the result and can exit without having to run the remaining layers~\cite{teerapittayanon2016branchynet, kaya2019shallow, fleet2023}. 
This is an orthogonal dimension to the idea of limiting the number of classes, so we augment $\mu$Classifiers with early exit capability. 
We compare the performance of a $\mu$Classifier with early-exit capability against an all-class models with early exit.

Figure~\ref{fig:early_exit accuracy_vs_size} shows the average accuracy across all 3-class subsets, 4-class subsets, and so on till all 10 classes after applying quantization and adding two early exit branches (stages 4 and 6) to EfficientNet-B0. 
We see that accuracy increases as the number of classes decreases. 
In particular, 3-class $\mu$Classifiers have 5\% and 8.6\% higher accuracy compared to all-class models for STL-10 and Enonkishu datasets respectively. 
Thus, the performance advantages of $\mu$Classifiers remain even if we augment the model with early exits.

%% file: DesignOverview.tex
\section{\myname{} Design Overview}
\label{des_over}

We build on these observations to design a context-aware inference pipeline \myname{}, which executes context-aware $\mu$Classifiers for a prediction task and dynamically switches between $\mu$Classifiers to handle context changes.

\begin{figure*}[t!!!]
    \centering
    \includegraphics[width=1\textwidth]{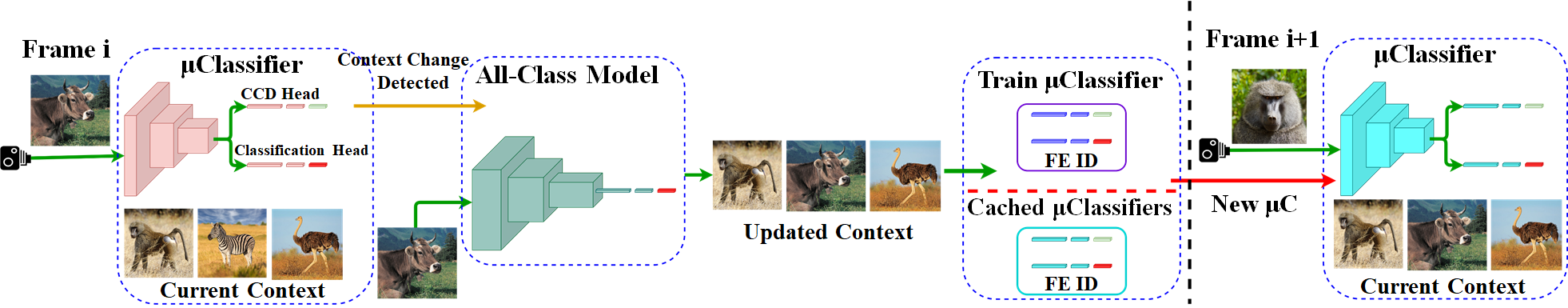}
    \caption{Context-Aware Switching in \myname{}. In case the communication or cloud is not available, the framework relies on cached/stored $\mu$Classifiers. (CCD: Context change detection, FE: Feature extractor)}
    \label{framework_all}
\end{figure*}


\myname{} has two main components.
The first component, \textit{Configuration Predictor}, identifies the suitable configuration of the $\mu$Classifier for each context, i.e., a subset of classes (called active classes), without any training costs.  
It represents a context based on an \textit{inter-class similarity} metric, which estimates the difficulty in classifying the active classes in a context. 
The metric allows a simple k-nearest neighbor approach to rapidly predict the optimal configuration of a $\mu$Classifier without training every possible configuration.

The second component, \textit{context-aware switching}, enables the device to seamlessly switch between $\mu$Classifiers based on context changes. 
It occurs in two steps. 
To detect context changes, we augment $\mu$Classifiers with a lightweight \textit{Context Change Detector} head. 
The first step executes the \textit{Context Change Detector} head to rapidly detect if the input frame is a new class without determining which is the new class. 
If context change is detected, the second step executes the \textit{all-class model}, a heavyweight classification module, that determines which classes are present in the scene. 
If a new class is detected, it triggers a ``model switch'' to load a new $\mu$Classifier based on the context switching policy. 
Figure~\ref{framework_all} illustrates switching at run-time.


\mypara{Usage scenarios:} \myname{} supports cloud-assisted and unattended operation. When cloud assistance is available, context-aware switching detects changes in context and requests a $\mu$Classifier for the new context. \myname{} cloud executes the Configuration Predictor to rapidly identify the suitable configuration of the $\mu$Classifier, then trains the model on-demand and sends the model to the device to handle the new context. The new model can be cached on-device and the next time the same context occurs, the cached model can be used as shown in Figure~\ref{framework_all}.
For unattended operation i.e. no cloud availability, \myname{} relies on locally stored $\mu$Classifiers. 
%
In variable connectivity settings, a combination of cloud-assisted and unattended operation is also possible. We now describe the two components in more detail.


%% file: Design.tex
\section{Configuration Predictor}
\label{sec:similarity}

\myname{} needs to train a resource-efficient $\mu$Classifier for every possible context. 
As the optimal $\mu$Classifier configuration for different contexts varies, the core challenge lies in how to quickly determine the optimal configuration of the $\mu$Classifier given a context.  

To see why the optimal $\mu$Classifier configuration varies across contexts, consider classifying animals captured by a trail camera. 
When a context consists of animals with similar features, textures, or behaviors, the $\mu$Classifier faces a more challenging task in distinguishing between them. 
For instance, if the three classes represent a deer, an elk, and a moose, the classifier has to contend with the fact that all three animals have somewhat similar body shapes, sizes, and features. Consequently, the model may need to employ more computational resources to scrutinize subtle distinctions and achieve a comparable level of accuracy.

To address the challenge, our idea is to design an ``inter-class similarity'' metric that captures the similarity between the set of classes in a $\mu$Classifier, which in turn correlates with how much computational resources (i.e., the configuration) the model requires.
Based on the metric, we design a lightweight kNN-based configuration selector to identify the best configuration of the $\mu$Classifier without needing to train all configurations. 
The configuration selector can be queried to output the best configuration i.e. the one that achieves the lowest resource demands while meeting a target accuracy for a set of classes (i.e., a context).

\subsection{Inter-Class Similarity Metric}

Given a set of classes, the inter-class similarity metric is a set of real numbers, each representing a \textit{pairwise class similarity}.
Two sets of classes that have similar statistics of inter-class similarities would share a similar $\mu$Classifier configuration. 
We use the statistics of inter-class similarities to represent a context, called \textit{context representation}. 

\mypara{Pairwise Class Similarity:}
To measure the similarity between two classes, we first need to find a representation for each class. 
For each input image of a specific class, we get its embedding from a feature extractor that is trained on all classes. 
Given a set of input images of that class, we can average the embedding of all images as the \textit{class representation}. 
The similarity between two classes is calculated as the Cosine Similarity between their class representations. 

\mypara{Context Representation:} The vector representation of a set of classes  (i.e., a context) is the mean and standard deviation of similarities among all class pairs.
Mathematically, let $s_{i, j}$ be the cosine similarity between class $i$ and class $j$. Then the representation of an $m$ class combination $C$ ($|C|=m$), is:
\begin{equation}
    [\text{mean}(\{s_{i, j}\}), \text{std}(\{s_{i, j}\})], i, j \in C. \label{eq:rep_vector}
\end{equation}

We note that the above class similarity and context representation were chosen after consideration of several alternate measures of similarity. 
We show empirical results comparing these metrics in \S\ref{sec:predictor}. 



\begin{figure}[h!!!]
    \centering
    \includegraphics[width=0.4\textwidth]{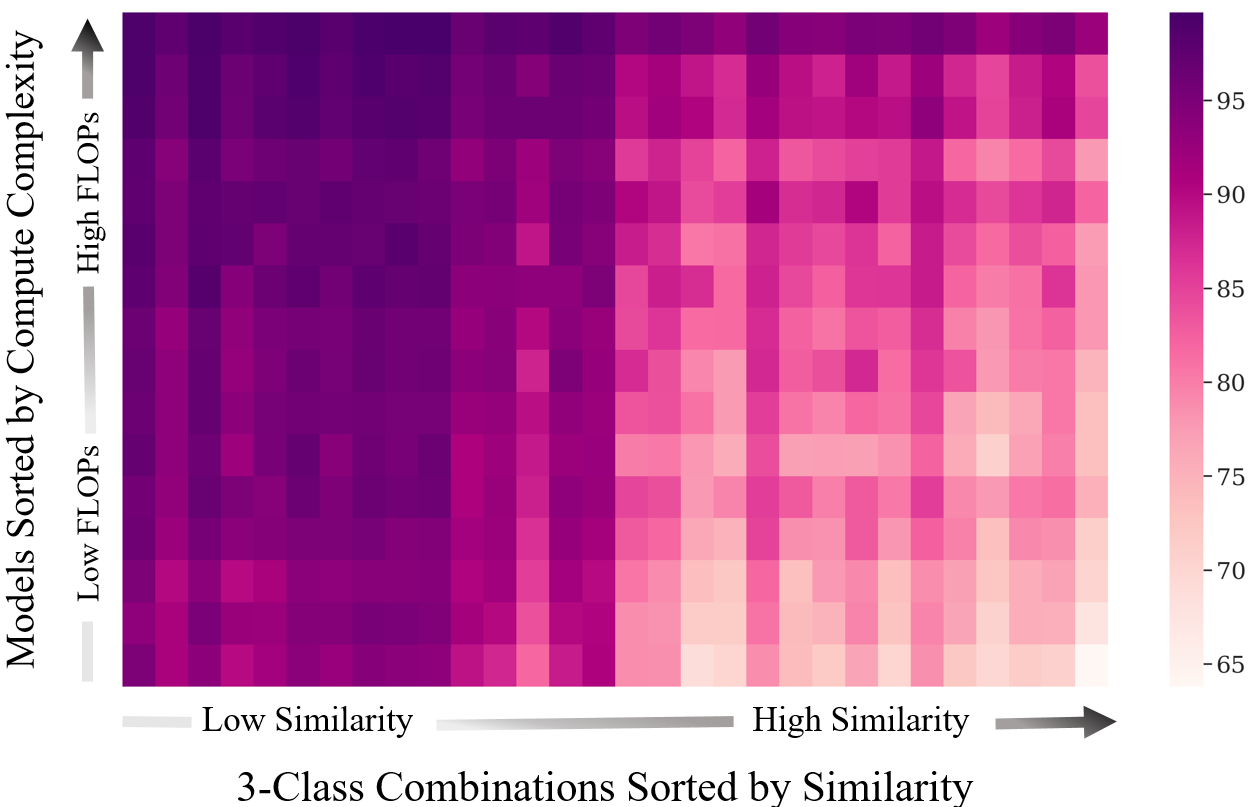}
    \caption{Model complexity versus the proposed inter-class similarity metric among classes (darker is higher accuracy).}
    \label{heatmap}
\end{figure}

Figure~\ref{heatmap} shows that the mean value of the inter-class similarity metric generally tracks the difficulty level of a context. 
 We sample 30 different 3-class combinations from the  STL-10 image classification dataset. We sort these combinations in terms of their mean similarity (Eq. \ref{eq:rep_vector}) from low to high (X-axis). 
The Y-axis shows the model complexity of $\mu$Classifiers measured by FLOPs. 
Each entry in the heatmap reports the accuracy of the $\mu$Classifier with a specific FLOPs for a 3-class combination (darker is higher accuracy). 
Overall, \textit{the higher the mean value of inter-class similarity, the less accurate a $\mu$Classifier with the same configuration (i.e., same FLOPS) since the context is more difficult. }
The figure indicates that our proposed inter-class similarity metric correlates well with the computational efforts required by a context.

\begin{figure}[t]
    \centering
    \includegraphics[width=0.45\textwidth]{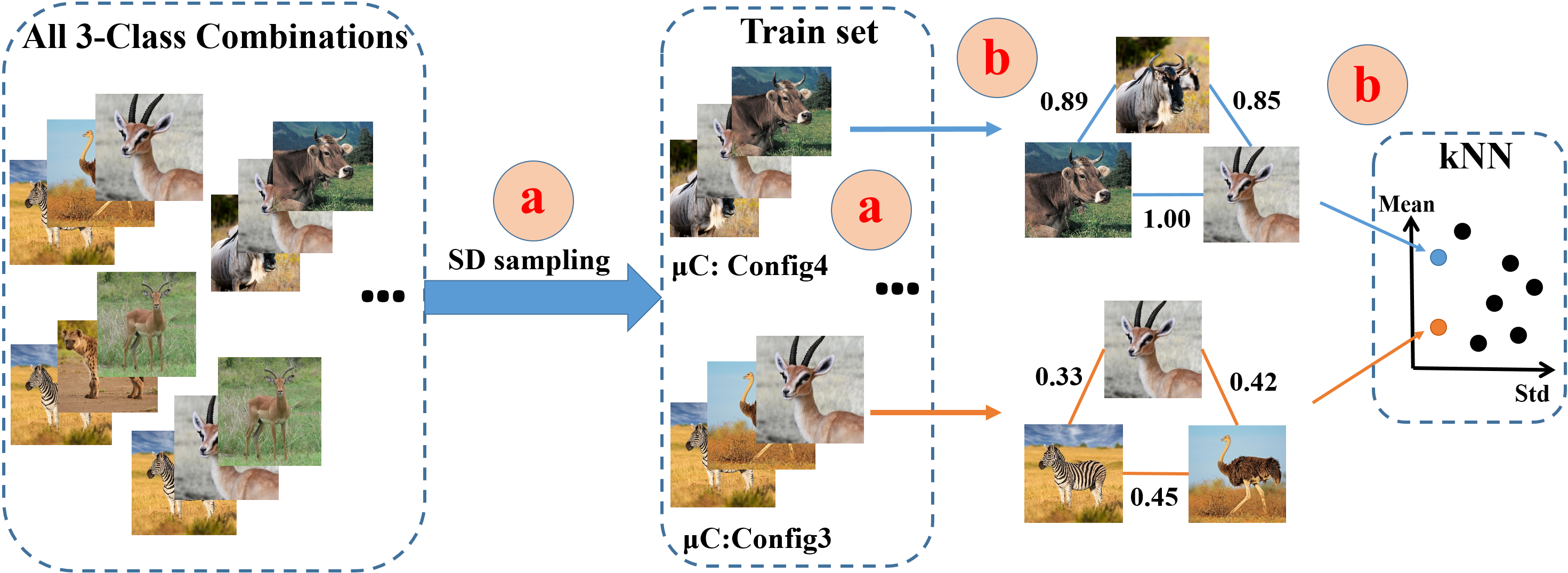}
    \caption{To build the training dataset for the kNN, we (a) use similarity-directed (SD) sampling to select $x$ subsets of classes, and determine the best configuration for each subset as the ground truth label (e.g., Config3 and Config4); and (b) for each subset, calculate the mean and std of inter-class similarities to get its context representation.}
    \label{fig:kNN-training}
    \vspace{-0.1in}
\end{figure}

\subsection{KNN-based Configuration Predictor}
\label{sec:knn}

The inter-class similarity metric allows us to develop a simple kNN-based approach to estimate the optimal configuration for a $\mu$Classifier. 
Building the kNN-based configuration predictor requires a configuration space for $\mu$Classifiers, and a set of training data points that consist of context representations and their corresponding optimal configurations.  

\mypara{Configuration Space:}
While in principle the configuration space can include any model compression method and its parameters, for practical reasons, we restrict ourselves to a few parameters to make training tractable. We therefore restrict our focus to configuration parameters that provide the large dynamic range of resource-accuracy tradeoffs for $\mu$Classifiers.
We find that two methods are particularly effective for configuring $\mu$Classifiers --- changing the input image resolution and the number of filters via filter pruning. 
In addition, we utilize post-training quantization to further lower the inference runtime and reduce the model size.

\mypara{Similarity-directed sampling:} 
The challenge of training the kNN in a resource-efficient manner is to build an effective training set. 
On one hand, we want samples to have good coverage of the spectrum of $m$-class combinations so that we can predict the best configuration of the remaining combinations accurately. 
On the other hand, we want as small amount of samples as possible because for each sample, we need to train all possible $\mu$classifiers (configurations) to identify the best one. 

To sample efficiently, we propose a {similarity-directed sampling} scheme that samples at different levels of hardness among all possible subsets of classes. 
Figure \ref{fig:kNN-training} illustrates the training of kNN.
First, we compute the mean of pair-wise similarity for all combinations. 
We then cluster them into four groups of similarity (quartiles). We then sample randomly from each cluster to gather enough data samples for training the kNN. 
Compared to random sampling, our empirical evaluation on three datasets shows that our sampling scheme can reduce up to 25-66\% of samples while getting similar prediction accuracy.

\mypara{Querying the kNN:} 
Once the kNN is constructed, it can be queried at run-time to predict the optimal $\mu$Classifier for the current context. 
The mean and std of inter-class similarities of the classes in the context are used to determine the closest configuration parameters in the kNN as the predicted optimal $\mu$Classifier configuration.
Feature extractors of $\mu$Classifiers are frozen during training to save storage overhead on device and also training time on the cloud. 



We note that an alternate approach to using a kNN-based configuration predictor is to use Neural Architecture Search (NAS \cite{NAS1,NAS2}). 
Our method is considerably lighter weight since, given a context, we can cheaply determine which configuration parameters to use via the kNN.

\section{Context-aware Switching}
\label{context_change}

Another essential part of making our system work is precisely detecting context change and, once detected, rapidly switching to the right $\mu$Classifier for the new context. 
 
This poses two challenges. 
The first challenge is to robustly and rapidly detect context changes. 
A context change occurs when an input frame falls into an unknown class that is not covered by the current $\mu$Classifier. 
Since context change detection lies on the critical path of every prediction, the detector should be very fast. 
The second challenge is to determine the appropriate $\mu$Classifier to switch to after the new class is identified.  Switching to a $\mu$Classifier with less number of classes can reduce the inference cost per input frame as the $\mu$Classifier can be more lightweight. However, it can introduce more frequent context switches and thus increase runtime overhead in the future. 
We now explain the proposed context change adaptation pipeline and the hybrid context switching policy that address the two challenges. 

\subsection{Context Change Adaptation Pipeline}

We design an efficient context change adaptation pipeline by splitting the process into two parts – a lightweight \textit{context change detector} that executes on each image
as a separate head attached to each $\mu$Classifer, and a more
heavyweight \textit{context predictor} that executes a regular all-class classifier to identify the new class. This separation
allows us to invoke the heavyweight classifier sparingly thereby reducing the computational cost.

\mypara{Context change detector:}
The goal of this module is to detect when the current context has changed. 
The problem is essentially the same as detecting out-of-distribution samples, which falls into the area of uncertainty estimation~\cite{ensemble1,mcdrop1,testtime1,testtime2,maxsoft1,naff1,naff2,aff1,aff2,aff3}. 
Although many approaches have been proposed for uncertainty estimation, most of them are computationally intensive and thus cannot be applied in a real-time scenario (see \S\ref{related}).  
Among all the approaches, Maximum Softmax Probability~\cite{maxsoft1} is the cheapest way to capture out-of-distribution samples. 
However, as we show later, this method yields unsatisfactory performance in our work due to high false positive and negative rates.

In this work, we introduce a regression-based context-change detector. 
Specifically, a regression head is added to the $\mu$Classifier as the second output head along with the classification head. 
The regressor outputs a value around 0 for the classes on which the $\mu$Classifier is trained and a value around 1 for all other classes i.e. for a new context. 
Both the classification and regression heads share the feature extractor of the $\mu$Classifier and thus the additional computation overhead introduced by the regressor is negligible. 
Since the output of the regressor is a continuous value, a threshold, $\theta$, is needed to distinguish between the current context and the new context. 
This threshold is dataset-dependent and tunable considering the trade-off between the ratio of false positives and false negatives. 
We set $\theta=0.5$ by default.

\mypara{Context predictor:}
After a context change is confirmed by the context change detector, the next step is to identify the new class. This is accomplished by an \emph{all-class} model that determines which are the classes that need to be included in the new $\mu$classifier. 
The \emph{all-class} model can be either executed locally or at the edge cloud depending on connectivity. Executing the \emph{all-class} model at the edge cloud allows us to use more powerful and more accurate models for determining the classes in the current context, which in turn can lead to more precise switching.

\subsection{Hybrid Context Switching} 
\label{sec:switch-policy}
After a new class is identified, the challenge is to decide what size $\mu$Classifier to switch to. The tradeoff is that using $\mu$Classifiers with a larger number of classes can reduce future context switches and improve accuracy whereas using  $\mu$Classifiers with a small number of classes can reduce inference latency per input frame. 
 
Our insight are two-fold. First, the right balance depends on the deployment environment: A device that encounters much more active classes than another device is likely to prefer $\mu$Classifiers with more classes to avoid higher context switch overheads.  
We introduce a configuration parameter that can be easily tuned based on the target deployment environment to determine how the $\mu$Classifiers are switched. 

\begin{figure}[h]
\includegraphics[width=0.35\textwidth]{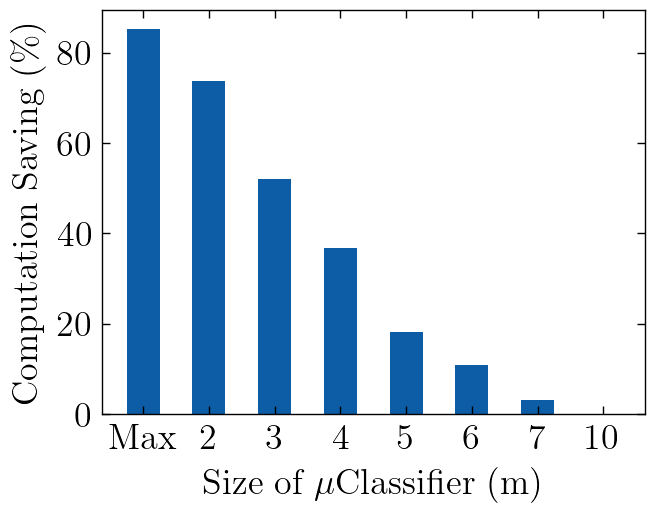}
    \caption{Computation saving vs $\mu$Classifier size ($m$).}
    \label{fig:comp_save_muc_size}
\end{figure}

Second, we do not need to consider all possible $\mu$Classifier sizes since the computational benefits of $\mu$Classifiers drops dramatically as the number of classes in them increases.  
Figure~\ref{fig:comp_save_muc_size} illustrates this using the PLACES20 dataset, which consists of a total of 20 classes from PLACES365 ~\cite{PLACES365}. 
For each combination of $m$ classes, we select the most efficient configuration that meets a pre-defined accuracy threshold. Then we calculate an average of the computation savings compared to the all-class model for all possible combinations of $m$ classes. On the X-axis, "Max" denotes the maximum computation savings, achieved by consistently using the lowest FLOPs configuration.
As the number of classes in a $\mu$Classifier goes beyond ten, the $\mu$Classifier brings no computation savings compared to the full-class model.
This motivates us to prioritize $\mu$Classifiers with a small number of classes (e.g., 2 to 5) to capitalize on the computation savings from context-aware inference. 

Based on the two insights, we develop a ``hybrid'' context switch policy that switches between $\mu$Classifers based on the deployment environment. 

 Our policy works as follows.
A kNN is already trained for each $m \in \{2, 3, 4, ...\}$ with a predetermined accuracy threshold.
We start by looking at the configuration needed for the m-class $\mu$Classifier (i.e. context size = m) where $m \in \{2, 3, 4, ...\}$. We then pick the context ($\mu$Classifier) with the most efficient configuration. If multiple context sizes share the same configuration, we select the largest context size that corresponds to this configuration. This allows us to pick a minimal $\mu$Classifier configuration while also minimizing future context switches by picking as large as possible context size.

We now explain two optimizations to efficiently implement the context switching policy in both cloud-assisted and unattended operation modes. 

\mypara{Caching $\mu$Classifiers:}
In cloud-assisted mode, a device downloads the $\mu$Classifier heads for a new context and caches them on device to avoid model training for the same context. 
If the cache is full due to device storage constraints, the least frequently used model will be replaced. 
Model caching reduces the overhead of cloud-side model training and device-cloud communication.

In unattended mode, a set of $\mu$Classifiers will be pre-installed on the device before deployment. 
The context switching policy will switch between only the pre-installed $\mu$Classifiers. 
We select the $\mu$Classifiers that are most frequently used based on an input trace collected from the IoT device deployed in a target environment.

\mypara{Optimizing storage requirements:} 
$\mu$Classifiers that operate on the same model size share the same model weights as their feature extractors and are frozen during training. We can reduce storage requirements by storing one feature extractor per model size, and only storing the classification and context change detection heads for each chosen $\mu$Classifier. 

Specifically, in cloud-assisted mode, a device stores all the unique sized feature extractors, one classification and context change detection head for the current context, and if storage is available, additional classification and context change detection heads from previous contexts. 

In the unattended mode (i.e., without cloud availability), a device stores all the unique sized feature extractors, an \emph{all-class} model for context predictor, and, subject to storage capacity, classification, and context change detection heads from chosen contexts based on the trace of the training data. 
These cached contexts are the most frequent when we apply the hybrid context-switching policy on the training set.
We show in \S~\ref{sec:case-study} that the total amount of storage required for 60 $\mu$Classifiers plus feature extractors and the \emph{all-class} model is only 28MB. 

%% file: Evaluation.tex
\section{Evaluation}
We first describe the experiment settings in \S~\ref{sec:settings}, and then evaluate the end-to-end system over five datasets and on real platforms  in \S~\ref{sec:end-to-end} and the components of \myname{} in \S~\ref{sec:predictor}-\ref{sec:selector}. 
We further perform ablation studies on context change frequency and scalability of the system in \S~\ref{sec:ablation} and implement a camera trap application in \S\ref{sec:case-study}.

\subsection{Experiment Settings}
\label{sec:settings}

\mypara{Dataset:}
\label{dataset}
We use five datasets. Three datasets come from the Camera Trap applications called Enonkishu \cite{Camera_trap_enonkishu}, Camdeboo \cite{Camera_trap_camdeboo}, and Serengeti (season 4) \cite{Camera_trap_serengeti}. 
We refer to them together as Camera Trap datasets. 
They have temporal data captured by trail cameras with real-world context changes. Trail cameras make use of an infrared motion detector and time-lapse modes to capture images, each time in a sequence of three. 
The Camera Trap datasets normally include a lot of species plus an empty class whose images show a scene with no species in it. 
Not all species have enough samples for both train and test sets. Hence we considered the most frequent species plus the empty class for the classification task. The number of considered classes were 9, 11, and 18 for Enonkishu, Camdeboo, and Serengeti respectively. 

We also evaluate on two other image datasets, STL-10 \cite{STL10}, and PLACES365 \cite{PLACES365}, from which we synthesize temporal sequences.
STL-10 is an image recognition dataset 
consisting of RGB images with a size of 96$\times$96. 
The dataset contains 10 classes, each with 1300 labeled images. 
PLACES365 is intended for visual understanding tasks such as scene context, object recognition, and action prediction.  
We chose 100 scene categories, each with 1200 samples of size 256$\times$256. 
While Most results are shown for 20 scene categories (called PLACE20), we evaluate \myname{}'s scalability for 100 classes.

\mypara{KNN-based configuration predictor:}
The $\mu$Classifier is based on the feature extractor of EfficientNet-B0~\cite{efficientnet} pre-trained on ImageNet. 
We vary both the pruning level and the resolution of input frames to get different configurations of the feature extractor.  
We applied three levels of $\ell1$ pruning, 10\%, 30\%, and 40\%, so that including the EfficientNet-B0, we have a total of four unique-sized feature extractors.  
The range of image resolutions is considered 100--320, 58--96, and 130--256 for Camera Trap, STL-10, and PLACES datasets respectively. 
Overall, the size of the configuration space is 16.
A $\mu$Classifier consists of one of the four feature extractors and two heads, one for classification and one for regression (to detect context change). 
Each head uses two Fully Connected layers. 
Feature extractors are fine-tuned once on all classes and remain frozen during the training of $\mu$Classifiers' heads.


To build the training data for the configuration predictor, we applied similarity-directed sampling (detailed in \S\ref{sec:knn}) on m-class combinations.
For each sampled $m$-class combination, we train $\mu$Classifiers with all 16 configurations to select the most efficient one that achieves a target accuracy (i.e., the optimal $\mu$Classifier configuration). 
We further use the feature extractor of ResNet-50 to extract the embedding of images from a class and average the embeddings into a single vector to compute context representation as Eq.~\ref{eq:rep_vector}.
The pairs of context representation for the sampled $m$-class combinations and the optimal $\mu$Classifier configuration for the context are the training dataset of the kNN. 


To balance computation efficiency and accuracy, the target accuracy threshold for selecting the best $\mu$Classifier is set to be around 94$\%$, 87$\%$, 92$\%$, 92$\%$, and 97$\%$ for Enonkishu, Camdeboo, Serengeti, STL-10, and PLACES20 datasets respectively, unless noted differently. 

After collecting the training data, we use the kNN to predict the best $\mu$Classifier configuration for the remaining $m$-class combinations. 
kNN predicts the configuration of a $\mu$Classifier based on the majority vote. It starts with 3 nearest neighbors and if there is no majority, it considers 4 nearest neighbors and so on until a majority is achieved.

\mypara{Context-aware switching:} 
As discussed in \S\ref{context_change}, a regression head is attached to a $\mu$Classifier to detect context change. 
Once context change is detected, we execute an all-class model (ResNet-50 for the edge-cloud assisted setting and EfficientNet-B0 for unattended op.) to identify which new class occurs in the scene. 
All-class models are pre-trained on ImageNet and fine-tuned on the target dataset.

\mypara{Implementation:}
We implement the \myname{} pipeline on three different IoT devices --- Raspberry Pi 4B, Raspberry Pi Zero W, and GAP8. 
We generate \textbf{TensorFlow Lite} models using the default optimization and int-8 quantization to reduce model size and execution latency. On all the platforms, we profile the execution latency of models of different configurations on different datasets.
For profiling on Raspberry Pi 4B and Zero W, we utilize the \textbf{TVM} framework \cite{TVM} to tune the operator implementations to reduce the execution latency for the Arm Cores. 
For profiling on GAP8, we use the GAP8 tool chain \cite{URL:GAP8_tool} to transform the \textbf{TensorFlow Lite} models into C code by fusing the operations. 
%
A cluster of NVIDIA Tesla M40 24GB GPUs was used for training models and one M40 GPU as the edge-cloud in some of our experiments. 
%
If accepted, we will make our code on all the above platforms openly available.

\subsection{End-to-End Evaluation}
\label{sec:end-to-end}

\begin{table*}[ht]
\small 
\tabcolsep=0.2cm 
\scalebox{1.0}{\begin{tabular}{@{}l|ccc|ccc|ccc|ccc@{}}
\toprule
Methods & \multicolumn{3}{c|}{Enonkishu} & \multicolumn{3}{c|}{Camdeboo} & \multicolumn{3}{c|}{Serengeti} & \multicolumn{3}{c}{PLACES20}\\
  &\multicolumn{1}{l|}{Acc} & \multicolumn{1}{c|}{Pi0} & GAP8 & \multicolumn{1}{c|}{Acc} & \multicolumn{1}{c|}{Pi0} & GAP8 & \multicolumn{1}{c|}{Acc} & \multicolumn{1}{c|}{Pi0} & GAP8 &\multicolumn{1}{c|}{Acc} & \multicolumn{1}{c|}{Pi0} & GAP8 \\
  \midrule
All-Class Model & 88.9\% & 1.0$\times$ & 1.0$\times$ & 65.4\% & 1.0$\times$ & 1.0$\times$ & 74.3\% & 1.0$\times$ & 1.0$\times$ & 90.5\% & 1.0$\times$& 1.0$\times$ \\
All-Class-Early Exit & 84.6\% & 1.7$\times$ & 2.0$\times$ & 65.5\% & 1.1$\times$ & 1.2$\times$ & 68.7\% & 1.5$\times$ & 1.5$\times$ & 90.1\% & 1.2$\times$& 1.3$\times$\\
All-Class-Pruned & 81.4\% & 1.2$\times$ & 1.7$\times$ & 57.7\% & 1.2$\times$ & 1.7$\times$ & 69.6\% & 1.2$\times$ & 1.7$\times$ & 84.9\% & 1.4$\times$& 2.0$\times$\\
FAST \cite{FAST} & 90.1\% & 1.1$\times$ & 1.2$\times$ & 64.9\% & 1.0$\times$ & 1.0$\times$ & 74.0\% & 1.0$\times$ & 1.0$\times$& 91.0\% & 1.0$\times$& 1.0$\times$\\
PALLEON \cite{palleon} & 90.7\% & 1.2$\times$ & 1.3$\times$ & 65.7\% & 1.0$\times$ & 1.0$\times$ & 74.1\% & 1.0$\times$ & 1.0$\times$ & 92.2\% & 1.0$\times$& 1.1$\times$\\\hline
\myname{} & \textbf{92.2\%} & \textbf{3.5$\times$} & \textbf{5.0$\times$} & \textbf{65.4\%} & \textbf{1.6$\times$} & \textbf{2.0$\times$} & \textbf{73.8\%} & \textbf{2.1$\times$} & \textbf{2.8$\times$} & \textbf{95.9\%} & \textbf{2.0$\times$} & \textbf{2.8$\times$}\\
\bottomrule 
\end{tabular}}
\caption{End-To-End performance of our approach vs baselines on three datasets; Context changes every 30 frames on average. For \myname{}, we used the hybrid switching policy with m-class contexts where $m \in \{2, 3, 4\}$. Speedup values are rounded. The gains improve if wireless data rates are higher (we use 3Mbps) or if caching is used (we assume no caching of the $\mu$Classifier heads). Due to space constraints, results for STL-10 are not included but they follow similar trends.}
\label{End-to-end-online}
\vspace{-0.2in}
\end{table*}

This section looks at an edge-cloud assisted system (the unattended setting is examined in \S~\ref{sec:selector}).
To evaluate the end-to-end performance of \myname{}, we need datasets that have context changes. The three Camera Trap datasets provide us with a real-world temporal flow with context changes but other datasets are not temporal. 
Hence, we sequence test sets from the PLACES and STL-10 datasets to mimic the temporal flow of the Camera Trap to create a temporal sequence with which to evaluate our methods. 
Specifically, we find that for Camera Trap datasets considering 3-class contexts, the average context interval is 30 frames (context changes in every 30 frames). 
We thus synthesize a test set with the context interval of 30 frames for the PLACES and STL-10 datasets and use this in our evaluation unless noted differently.
In \S~\ref{sec:ablation}, we further study the effects of different context change rates in the performance of \myname{}.


We compare the classification accuracy and overall speedup of \myname{} against the following baselines: 
(a) \textbf{FAST} \cite{FAST}, a context-aware approach that assumes the class skews are known offline. During inference it detects the class skew using a window-based detector and train a specialized classifier (top layers), 
(b) \textbf{PALLEON} \cite{palleon}, a state-of-the-art video processing framework that applies a Bayesian filter to adapt the model to the context. 
The Bayesian filter is determined by applying an all-class model on a window of frames.
(c) \textbf{All-Class Model}, which uses the well-trained EfficientNet-B0 with all the classes and full image resolution. 
(d) \textbf{All-Class-Early Exit}, which is the early exit variant (exit heads in stages 4 and 6)  of the All-Class Model.
(e) \textbf{All-Class-Pruned}, where we apply 30$\%$ pruning to the all-class model. We chose 30\% as it gives the best accuracy and efficiency trade-offs. 


Table~\ref{End-to-end-online} shows the end-to-end results on two devices. 
The IoT device communicates at a data rate of $3$ Mbps (typical LTE speed).
Computed latencies include the time spent on the inference on the device, transmission of the triggered frames to the cloud, and transmission of the classification and regression heads to the device. 
We assumed no $\mu$Classifiers are cached on the device which is the \emph{worst-case scenario} as it requires more transmission and incurs higher latency. We averaged latencies over the test set and computed the speedup relative to the All-Class Model. Note that for the Camera Trap datasets, accuracies are generally low regardless of whether we use EfficientNet or ResNet-50 due to data cases that are difficult to classify as well as inaccurate labels.



Overall, our approach has the advantage in terms of accuracy and latency over baselines. 
When compared to All-Class Model, 
\myname{} has up to 5.4\% higher accuracy and also offers latency speedup of $1.6 - 3.5 \times$ for the Pi0 and $2.0 - 5.0 \times$ for the GAP8.
%
\myname{} outperforms two other context-agnostic baselines, All-Class-Early Exit, and All-Class-Pruned, in terms of both accuracy and latency by leveraging context awareness to improve efficiency. 

Compared to the context-aware approaches FAST \cite{FAST} and PALLEON \cite{palleon}, \myname{} demonstrates comparable accuracy on the Camera trap datasets, and 3.7$\%$ -- 4.9$\%$ accuracy improvement for the PLACES20 dataset, but has 1.6$\times$ -- 5.0$\times$ speedup. FAST and PALLEON need to execute the All-Class model for several frames to determine the class skew. Thus, they provide speedups only if the context change is very infrequent.
For this reason, their latency speedup is at most 1.3$\times$ which is significantly lower than \myname{}.

\subsection{Performance of Config Predictor}
\label{sec:predictor}

We now look at the performance of the $\mu$Classifier \textit{Configuration Predictor}. 
We vary the target accuracy threshold from the lowest to the highest of all the trained $\mu$Classifiers to show how the threshold affects the performance of the configuration predictor. 
A lower target accuracy will result in lighter $\mu$Classifiers and thus less FLOPs.

\begin{figure}[t]%
    \centering
    \subfloat[\centering Enonkishu\label{fig:CameraTrap_roc}]{{\includegraphics[width=.24\textwidth]{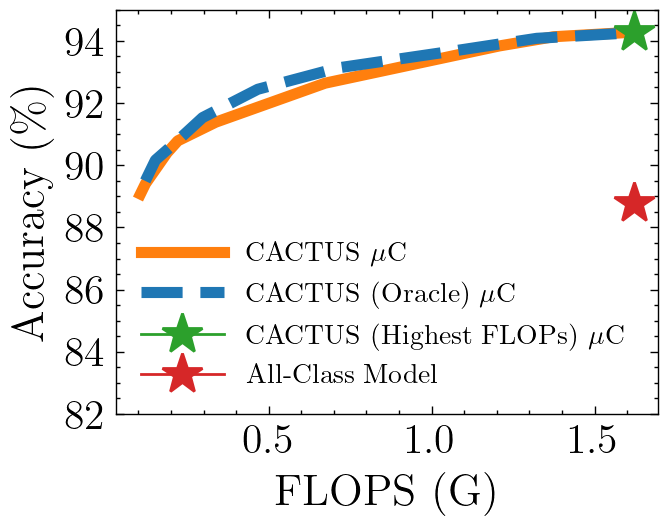}}}%
    \subfloat[\centering PLACES20\label{fig:Places_roc}]{{\includegraphics[width=.24\textwidth]{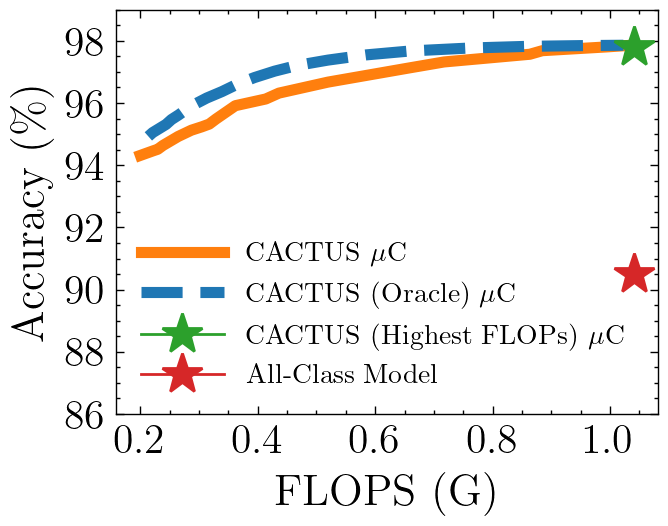}}}%
    \caption{Performance of configuration predictor with different accuracy thresholds.}
\label{knn_performance}
\end{figure}

Figure~\ref{knn_performance} shows how different target accuracy thresholds affect the classification accuracy and the computation cost for our method and baselines. 
We compare the \myname{}'s configuration predictor (noted as \textbf{\myname{} $\mu$C} in the figure) with three baselines: 
(a) \myname{} \textbf{(Oracle) $\mu$C} i.e. the $\mu$Classifier that meets the target accuracy threshold with minimum FLOPs (which would be selected by an omniscient scheme rather than the kNN-based method we use). 
We note that training and storing the Oracle is not scalable --- For example, using a 40 GPU cluster for training 3-class $\mu$Cs on the PLACES dataset, our method (similarity-directed sampling) is faster by 2 hours for 20 classes and 270 hours for 100 classes.
(b) \myname{} \textbf{(Highest FLOPs) $\mu$C} employs the $\mu$Classifier with the same classes as \myname{} but at full image resolution and without pruning (and hence computationally the most expensive).
(c) \textbf{All-Class Model}, which uses the well-trained EfficientNet-B0 with all
the classes and full image resolution.
We omit results for other datasets for limited space. 
Both the classification accuracy and FLOPs are averaged over $\mu$Classifiers for all 3-class combinations.

Overall, we see that \myname{} can cover a wide trade-off region between accuracy and computational cost --- accuracies vary by 3.5\% to 5\% and compute by 6.5$\times$ to 13$\times$. 
Across the range, the accuracy is nearly as good as the \myname{} (Oracle), indicating that \textit{the kNN together with the inter-class similarity metric works well in selecting the best configuration. }
In contrast, All-Class Model (red star) and \myname{} (Highest FLOPs) (green star) are point solutions. 
We see that \myname{} (Highest FLOPs) has significantly higher accuracy than the All-Class Model even though both models have the same FLOPs. 

\begin{figure}[t]%
    \centering
    \subfloat[\centering \label{fig:acc_vs_sim_eval}]{{\includegraphics[width=.24\textwidth]{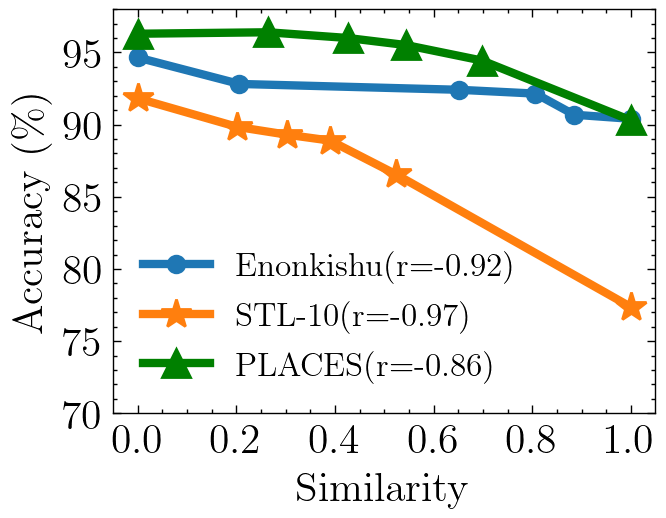}}}%
    \subfloat[\centering \label{fig:comp_gain_Vs_Similarity}]{{\includegraphics[width=.24\textwidth]{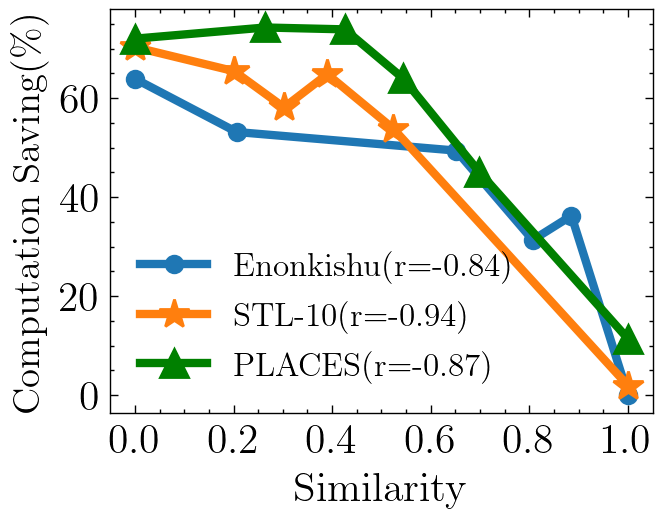}}}%
    \caption{(a) The accuracy of a representative configuration correlates well with the inter-class similarity. 
    (b) The computational complexity of the predicted $\mu$Classifiers correlates well with the inter-class similarity. 
    Similarity is normalized to [0, 1] in both figures.}
    \vspace{-0.2in}
\end{figure}



\mypara{Effectiveness of the inter-class similarity metric:}
The good predictive performance of the configuration predictor results from the effectiveness of the inter-class similarity metric in capturing the difficulty level of contexts. 


Figure~\ref{fig:acc_vs_sim_eval} illustrates how the accuracy of a representative $\mu$Classifier on each 3-class combination correlates with the inter-class similarity averaged over the 3 classes (Eq. \ref{eq:rep_vector}). 
For the graph, all 3-class combinations were binned into six clusters based on the inter-class similarity metric and the accuracy and similarity values are averaged in each cluster.
We observed that there is a clear inverse relation between inter-class similarity and classification accuracy. The Pearson correlation between the inter-class similarity metric and the classification accuracy is extremely high ($0.86-0.97$), indicating that \textit{our similarity metric closely mirrors the difficulty level of 3-class combinations.}


Figure~\ref{fig:comp_gain_Vs_Similarity} looks at this relation from another performance dimension --- computation savings. 
The same procedure of grouping 3-class combinations into six bins is applied here.  
The Y-axis shows the computation saving of the predicted configurations (by kNN) compared to the All-Class Model/ \myname{} (Highest FLOPs) and the X-axis is the inter-class similarity. 
We see the computation saving drops as the similarity increases since a more powerful model should be used to meet the target accuracy threshold. 
It also shows that the similarity-aware kNN works well in selecting lightweight configurations for low similarity combinations and more powerful configurations for harder cases.

\begin{figure}[h]
    \centering
\includegraphics[width=0.35\textwidth]{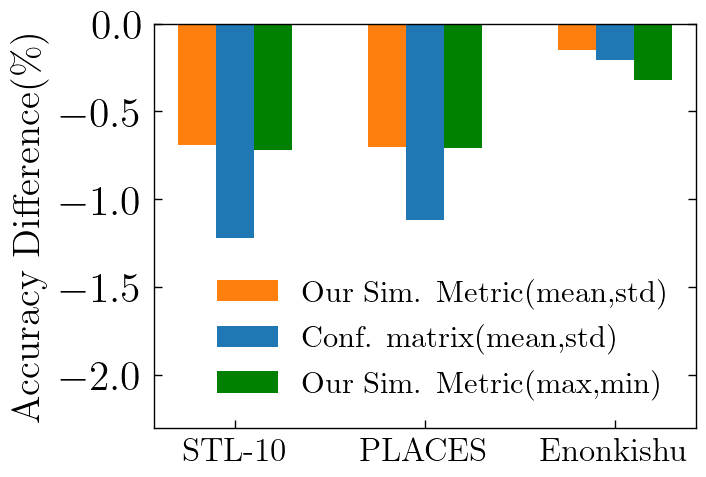}
    \caption{Accuracy difference of predicted $\mu$Cs by kNN from oracle $\mu$Cs using different similarity metrics (Predicted-Oracle).}
    \label{fig:Diff_Sim_Metric}
    \vspace{-0.1in}
\end{figure}

\mypara{Comparison with other similarity metrics:}
We also investigate alternative similarity metrics and context representations including (1) confusion matrix-based similarity and (2) maximum and minimum of pair-wise similarities in a combination. 
For computing the pair-wise similarity using the confusion matrix, the number of samples that two classes are misclassified as another are summed and normalized as the percentage by the total number of those two classes.
Figure~\ref{fig:Diff_Sim_Metric} shows the comparison using three datasets. Each bar reflects the accuracy difference between predicted $\mu$Classifiers and oracle $\mu$Classifiers averaged over all 3-class combinations for each similarity metric. We see kNN using our similarity metric (mean,std) provides more accurate predictions in comparison to the confusion matrix-based similarity metric over the three datasets. 

\subsection{Perf. of Context-Aware Switching}
\label{sec:switching} 

This section evaluates the performance of the \textit{context-aware switching} module. We first evaluate the performance of the regression-based context change detector and then the hybrid context switching policy. 

\mypara{Effectiveness of the context change detector:}
A false positive (FP) that wrongly predicts a context change would introduce additional cost whereas a false negative (FN) will cause accuracy degradation. 
To evaluate the performance of the regression head, we test the $\mu$Classifier corresponding to a 3-class combination with the test samples of 3 classes to calculate the FP rate. In order to compute the FN rate, we test the mentioned $\mu$Classifier with samples of all out-of-context classes. For each approach, the reported FP and FN rates are averaged over all combinations of 3 classes.

\begin{table}[htb]
\footnotesize 
\tabcolsep=0.05cm
\begin{tabular}{l|cc|cc|cc@{}}
\toprule
\multirow{2}{*}{\begin{tabular}[c]{@{}l@{}}Method\end{tabular}} & \multicolumn{2}{c|}{Enonkishu} & \multicolumn{2}{c|}{STL-10} & \multicolumn{2}{c}{PLACES} \\
 & FP & \multicolumn{1}{c|}{FN} & FP & \multicolumn{1}{c|}{FN} & FP & \multicolumn{1}{c}{FN} \\ \midrule
$\mu$C+regression (ours) & 3.88\% & 13.61\% & 13.58\% & 10.91\% & 8.91\% & 8.18\% \\
$\mu$C+maximum prob~\cite{maxsoft1} & 16.96\% & 22.89\% & 31.44\% & 33.25\% & 19.49\% & 18.29\% \\  
Oracle $\mu$C+regression & 4.01\% & 13.73\% & 13.83\% & 11.15\% & 8.57\% & 7.64\% \\
\bottomrule
\end{tabular}
\caption{False Positive and False Negative rates of our approach vs baselines.}
\label{fpfnt}
\vspace{-0.2in}
\end{table}

Table~\ref{fpfnt} compares the context-change detection performance of \myname{} (``$\mu$C + regression'') against baselines: 
(a) the same $\mu$Classifier as \myname{} but with a different approach for context change detection based on maximum softmax probability \cite{maxsoft1}.
(b) the \myname{} (Oracle) $\mu$Classifier with regression head (``Oracle $\mu$C + regression''). 
%
\myname{} is much more accurate at context change detection compared to the baseline that uses the maximum softmax probability. 
The FP and FN rates of our approach are $10-20\%$ lower than the baseline. 
\myname{} is quite close to the Oracle $\mu$Classifer as well. 
We note that the FP and FN rates can be lowered by increasing the accuracy threshold explained in \S\ref{sec:settings} or by using larger $\mu$Classifiers (e.g. 4 or 5-class) which would incur fewer context switches. Also, FP and FN rates can be tuned via the context change detection threshold.

\mypara{Effectiveness of hybrid context switching:}
We now compare hybrid context switching (i.e. switching between 2-class, 3-class, 4-class, and 5-class contexts) against a fixed-class switching policy (e.g. only 2-class or 3-class or 4-class context). Table \ref{hybrid_perf} shows the accuracy and overall speedup for the Serengeti dataset. We see that for fixed-class switching, accuracy increases as the number of classes increases since the model is less likely to make mistakes when deciding whether to trigger the all-class model. But speedup drops since the models chosen are larger. \myname{} (Hybrid) gets the best of both worlds --- by intelligently switching, it achieves high accuracy but with more speedup than the fixed-class models.  Table~\ref{hybrid_perf} also shows that the performance of \myname{} (Hybrid) generally improves as the number of context sizes i.e. $m$ increases from $m \in \{2\}$ to $m \in \{2,3,4,5\}$. But we get diminishing returns after $m \in \{2, 3, 4\}$, so that is our sweet spot. 



\begin{table}[t]
\scalebox{0.88}{\begin{tabular}{@{}l|ccc@{}}
\toprule
Methods & \multicolumn{3}{c}{Serengeti}\\
  &\multicolumn{1}{l|}{Acc} & \multicolumn{1}{c|}{SpeedUp(Pi0)} & \multicolumn{1}{c}{SpeedUp(GAP8)} \\
  \midrule

\myname{}-Hybrid (2,3)& 73.6\% & 1.9$\times$ & 2.5$\times$\\
\myname{}-Hybrid (2,3,4)& 73.8\% & 2.1$\times$ & 2.8$\times$\\
\myname{}-Hybrid (2,3,4,5)& 73.8\% & 2.1$\times$ & 2.9$\times$\\
\myname{} (2-Class) & 72.8\% & 1.4$\times$ & 1.7$\times$\\ 
\myname{} (3-Class) & 74.1\% & 1.3$\times$ & 1.4$\times$\\  
\myname{} (4-Class) & 74.9\% & 1.0$\times$ & 1.1$\times$\\
\bottomrule
\end{tabular}}
\caption{Comparison between the hybrid switching policy versus fixed-class switching}
\label{hybrid_perf}
\vspace{-0.2in}
\end{table}

\subsection{Perf. of Local-Only \myname{}} 
\label{sec:selector}

We now consider unattended operation (i.e. local-only) --- this is the scenario where communication to the cloud is not available and the IoT device uses only locally cached $\mu$Classifiers. 
Since the all-class model also needs to be cached on the IoT device, we choose All-Class EfficientNet for this purpose since it is resource optimized. 

In order to decide which $\mu$Classifiers should be stored on the device, we used the training set and cached the most frequently used $\mu$Classifiers on the device. For local switching, if none of the chosen m-class $\mu$Classifier ($m \in \{2, 3, 4\}$) is available on-device, we simply select an available one that has the highest overlap in classes with the chosen one.

Table \ref{local_result} shows the accuracy and speedup on GAP8 for \myname{} and baselines. We evaluated \myname{} with 15, 20, and 30 stored $\mu$Classifiers. Results show that by only storing 30 $\mu$Classifiers, \myname{} can achieve 4.1$\times$ and 2.4$\times$ speedup for Enonkishu and Serengeti datasets respectively while the accuracy is on a par with baselines. Note that 30 $\mu$Classifiers constitute less than 1 percent of all possible 2, 3, and 4-class $\mu$Classifiers for the Serengeti dataset. Thus, even with a limited number of $\mu$Classifiers and storage \myname{} can maintain its advantage in speedup without sacrificing accuracy.

\begin{table}[t]
\scalebox{0.87}{\begin{tabular}{@{}l|c|cc|cc@{}}
\toprule
Methods & {} & \multicolumn{2}{c|}{Enonkishu} & \multicolumn{2}{c}{Serengeti}\\
  &\multicolumn{1}{l|}{Storage} &\multicolumn{1}{l|}{Acc} & \multicolumn{1}{c|}{SpeedUp} &\multicolumn{1}{c|}{Acc} & \multicolumn{1}{c}{SpeedUp} \\
  \midrule
All-Class Model& 4.8MB & 88.9\% & 1.0$\times$ & 74.3\% & 1.0$\times$\\
All-Class-Early Exit& 4.9MB & 84.6\% & 2.0$\times$ & 68.7\% & 1.5$\times$\\ 
All-Class-Pruned& 2.6MB & 81.4\% & 1.7$\times$ & 69.6\% & 1.7$\times$\\  
\myname{} (15 $\mu$Cs)& 19.9MB & 89.5\% & 2.2$\times$ & 73.6\% & 2.0$\times$\\
\myname{} (20 $\mu$Cs)& 20.8MB & 90.1\% & 3.5$\times$ & 73.9\% & 2.2$\times$\\
\myname{} (30 $\mu$Cs)& 23.3MB & 90.2\% & 4.1$\times$ & 74.0\% & 2.4$\times$\\
\bottomrule
\end{tabular}}
\caption{End-to-End performance of Local-Only \myname{} and baselines. Speedup numbers are computed for GAP8 (Rasp. Pi0 results show a similar trend).}
\label{local_result}
\vspace{-0.2in}
\end{table}

\subsection{Ablation Study}
\label{sec:ablation}
\mypara{Effect of context change rate:}
We also evaluated how our end-to-end results change as the interval between context change varies from every 60 frames to 4 frames on the GAP8 and Pi0 platforms using the PLACES20 dataset. 
Table~\ref{diff_context} shows that \myname{} maintains its advantage even with higher context change rates i.e. shorter context intervals. 
The speedup decreases slightly from 2.61--1.96$\times$ to 1.97--1.07$\times$ and accuracy gain decreases from 5.1$\%$ to 1.4$\%$ as the interval becomes smaller but \myname{} is still considerably better than the baseline (All-Class Model).

\begin{table}[t]
\scalebox{0.93}{\begin{tabular}{@{}l|lccccc@{}}
\toprule
\multirow{2}{*}{\textbf{Metrics}} & \multicolumn{6}{c}{Context Interval} \\
 & 4 & 5 & 10 & 20 & 30 & 60\\ \midrule
Accuracy Gain & 1.4\% & 2.3\% & 4.1\% & 5.0\%& 5.4\%& 5.1\% \\
Speedup (Pi0) & 1.07$\times$& 1.18$\times$& 1.56$\times$& 1.85$\times$& 1.96$\times$& 1.96$\times$\\ 
Speedup (GAP8) & 1.97$\times$& 2.01$\times$& 2.59$\times$& 2.83$\times$& 2.81$\times$& 2.61$\times$\\  
\bottomrule
\end{tabular}}
\caption{\myname{} accuracy gain and speedup for the GAP8 with different number of context intervals in comparison to the All-Class Model on PLACES dataset.}
\label{diff_context}
\vspace{-0.2in}
\end{table}

\mypara{Scalability:}
We now look at the scalability of \myname{} with increasing the number of classes. 
Table~\ref{scalability} shows how \myname{} end-to-end performance changes relative to the All-Class Model. We used a fixed 3-class switching policy since it was computationally intensive to train 2, 3, and 4-class $\mu$Classifiers for large number of classes. 
We see that \myname{}'s accuracy gain over the All-Class Model increases as we scale up the number classes, and its speedup reduces but it is still better. We expect higher speedup if the hybrid policy is utilized. We also see the benefits of similarity-directed sampling with more classes. For example, we only need 5\% of the $\mu$Classifiers to train the kNN when the number of classes is higher than 60.

\begin{table}[t]
\tabcolsep=0.07cm
\scalebox{0.93}{\begin{tabular}{@{}l|lccccccc@{}}
\toprule
\multirow{2}{*}{\textbf{Metrics}} & \multicolumn{7}{c}{Number of Classes} \\
 & 10 & 20 & 30 & 40 & 60 & 70 & 100 \\ \midrule
Accuracy Gain& 4.4\%& 4.7\%& 4.3\%& 4.6\% & 5.3$\%$ & 5.6$\%$ & 6.5$\%$ \\
SpeedUp (GAP8)&2.33$\times$& 2.26$\times$& 2.18$\times$& 2.10$\times$ & 1.96$\times$& 1.86$\times$ & 1.63$\times$\\ 
$SD_{sampling}$ ratio& 50.0$\%$& 17.5$\%$& 10.0$\%$& 7.5$\%$ & 6.0$\%$ & 5.0$\%$ & 5.0$\%$\\ 
\bottomrule
\end{tabular}}
\caption{Effect of scaling the number of covered classes on \myname{}'s performance (PLACES dataset). $SD_{sampling}$ is the ratio of $\mu$Cs used for training the kNN.}
\label{scalability}
\vspace{-0.2in}
\end{table}


\mypara{Overhead of model training:} We briefly report the training overhead for classification and context change detection heads. Since the feature extractor is pre-trained offline, we only need to train the $\mu$Classifier's heads which is very fast. The training overhead for Enonkishu, Camdeboo, Serengeti, and PLACES datasets are on average 5.4, 5.5, 5.7, and 3 seconds respectively on a Tesla M40 24GB GPU.

\subsection{Case study: Trail camera}
\label{sec:case-study}
We have developed an end-to-end implementation of a wild-life camera that can be deployed in the wild to capture images of animals or birds and classify them in real-time on the Raspberry Pi 4B. A demo that
shows a working end-to-end system is prepared (link\protect\footnote{ \protect\url{https://anonymous.4open.science/r/CACTUS_-5BF9/README.md}}).
In this implementation, we store $60$ $\mu$Classifiers heads and a total of four feature extractors. The total memory footprint of the models is $28$ MBs  and that of the feature extractors and all-class model is $17.6$ MBs. 
We use a sequence of $2100$ images from the Camera Trap dataset Enonkishu~\cite{Camera_trap_enonkishu} to visualize the demo but we can also directly obtain input from the Raspberry Pi camera. 

%% file: Related.tex
\section{Related Work}
\label{related}



\mypara{Deep Learning in low-power devices:}
Two popular approaches for the execution of Deep Neural Networks (DNNs) on resource-constrained IoT devices are early exit~\cite{teerapittayanon2016branchynet, kaya2019shallow} and model Compression techniques such as quantization \cite{review_comp,han2015deep, quant_app2}, model pruning \cite{review_comp,jacob2018quantization, li2020train, prune_app1}, and knowledge distillation \cite{hinton2015distilling, li2014learning}. 
These techniques are generic and usually can be combined with other approaches. In this work, we show that both model compression and early exit can be applied to $\mu$Classifiers to improve the computation efficiency. 


Another line of work is partitioned execution, where the IoT device executes a few layers of a model and offloads the remaining layers to the cloud. 
Some approaches \cite{kang2017neurosurgeon,teerapittayanon2017distributed} focus on locating the best partitioning point while others \cite{choi2018deep, cohen2020lightweight, huang2020clio, itahara2021packet} pay attention to intermediate features. However, this approach intrinsically requires continuous wireless communication that limits it to some scenarios.


\mypara{Efficient video processing:}
Exploiting the temporal locality of videos is the complementary direction for efficient IoT inference. Prior works fall into two categories. The first set of approaches \cite{Focus, Noscope, Chameleon} are query-based and assume scenarios are known offline (static context). The second category of approaches \cite{FAST,han2016mcdnn,palleon} consider dynamic context and try to adapt the deployed models. However, \cite{FAST} assumes class skews are known and \cite{han2016mcdnn,palleon} require network connectivity which limits them to specific scenarios. Also, \cite{FAST,palleon} are only effective for applications with long context intervals.

\mypara{Uncertainty estimation:} 
Our work is also related to uncertainty estimation since detecting context changes is essentially estimating the prediction uncertainty.  
There are multiple approaches for uncertainty estimation such as Ensemble methods~\cite{ensemble1}, Bayesian Approximations~\cite{mcdrop1}, Maximum Softmax Probability~\cite{maxsoft1}, Test-Time Augmentation~\cite{testtime1, testtime2}, and other deterministic methods \cite{uncertainty_review, naff1,naff2, aff1,aff2,aff3}.
Among these, Maximum Softmax Probability is the cheapest way to capture out-of-distribution samples.
The maximum value among outputted softmax probabilities for in-distribution samples tends to be larger than for out-of-distribution samples. This approach is inexpensive in terms of computation and memory as it doesn't need any specialized model and it only uses the classification softmax probabilities. However, it did not perform well in our experiments, hence we introduced a regression-based detector to meet our needs.

%% file: Conclusion.tex
\section{Conclusion}

In conclusion, our work presents a new paradigm, switchable $\mu$Classifiers, to the growing set of approaches for squeezing deep learning models on resource-constrained platforms. We showed that this method can improve accuracy while significantly lowering latency and therefore power consumption. Our work opens up a new direction and can spur follow-on work to fully explore the design space. While we focused on IoT devices in this work, context-aware inference may have broader applicability as models get larger and more complex, for example, in self-driving vehicles.